\documentclass[letterpaper]{article} 
\usepackage{aaai25}  
\usepackage{times}  
\usepackage{helvet}  
\usepackage{courier}  
\usepackage[hyphens]{url}  
\usepackage{graphicx} 
\usepackage{natbib}  
\usepackage{caption} 
\usepackage{algorithm}
\usepackage{algorithmic}

\usepackage{newfloat}
\usepackage{listings}
\DeclareCaptionStyle{ruled}{labelfont=normalfont,labelsep=colon,strut=off} 
\floatstyle{ruled}
\newfloat{listing}{tb}{lst}{}
\floatname{listing}{Listing}
\usepackage[export]{adjustbox}
\usepackage{bm} 
\usepackage{amssymb} 

\newcommand{\eg}{{e}.{g}., }

\usepackage{amsmath} 
\usepackage{subcaption} 
\usepackage{pifont} 
\usepackage{makecell}
\usepackage[table]{xcolor}
\usepackage{multirow}

\title{Two-stream Beats One-stream: Asymmetric Siamese Network \\
	for Efficient Visual Tracking}

\author{
    Jiawen Zhu\textsuperscript{\rm 1},
    Huayi Tang\textsuperscript{\rm 2}, Xin Chen\textsuperscript{\rm 1}, Xinying Wang\textsuperscript{\rm 1}, Dong Wang\textsuperscript{\rm 1}, Huchuan Lu\textsuperscript{\rm 1}\thanks{Corresponding author.}
}
\affiliations{
    \textsuperscript{\rm 1}Dalian University of Technology, Dalian, China\\

    \textsuperscript{\rm 2}University of Pennsylvania, Philadelphia, USA\\
    jiawen@mail.dlut.edu.cn, huayit@seas.upenn.edu \\
    \{chenxin3131, wangxinying\}@mail.dlut.edu.cn,
    \{wdice, lhchuan\}@dlut.edu.cn
}

\usepackage{bibentry}

\begin{document}

\maketitle

\begin{abstract} 
	Efficient tracking has garnered attention for its ability to operate on resource-constrained platforms for real-world deployment beyond desktop GPUs.
	Current efficient trackers mainly follow precision-oriented trackers, adopting a one-stream framework with lightweight modules.
	However, blindly adhering to the one-stream paradigm may not be optimal, as incorporating template computation in every frame leads to redundancy, and pervasive semantic interaction between template and search region places stress on edge devices.
	In this work, we propose a novel asymmetric Siamese tracker named \textbf{AsymTrack} for efficient tracking.
	AsymTrack disentangles template and search streams into separate branches, with template computing only once during initialization to generate modulation signals.
	Building on this architecture, we devise an efficient template modulation mechanism to unidirectional inject crucial cues into the search features, and design an object perception enhancement module that integrates abstract semantics and local details to overcome the limited representation in lightweight tracker.
	Extensive experiments demonstrate that AsymTrack offers superior speed-precision trade-offs across different platforms compared to the current state-of-the-arts.
	For instance, AsymTrack-T achieves 60.8\% AUC on LaSOT and 224/81/84 FPS on GPU/CPU/AGX, surpassing HiT-Tiny by 6.0\% AUC with higher speeds.
	The code is available at \textit{https://github.com/jiawen-zhu/AsymTrack}.
\end{abstract}

\section{Introduction}

As a long-standing fundamental topic, visual tracking aims at pinpointing the position of a target object in video frames.
Promising advancements have been achieved, attributable to increasingly powerful designs of deep models~\cite{resnet,attention_is_all,vit,diao2024gssf}.
However, in practical application scenarios, current high-performance trackers~\cite{ostrack, mixformer, artrack, vipt,chen2024masktrack,lorat} often fail to meet the low computational latency requirements, particularly on resource-constrained platforms.
Thus, designing efficient tracker is critical and recently attracts extensive research in industry and academia.

\begin{figure}[t]
	\centering
	\includegraphics[width=0.97\linewidth]{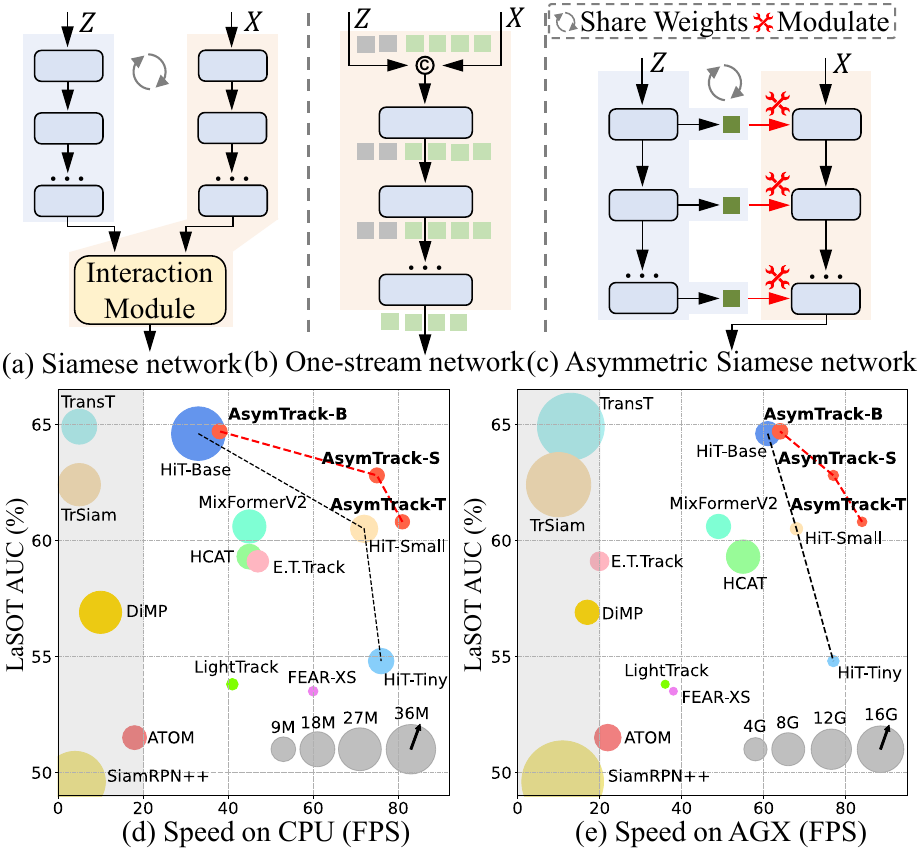}
	\caption{
	AsymTrack vs. other frameworks and trackers. (a)-(c) represent Siamese (two-stream) network, one-stream network and our asymmetric Siamese network, respectively. \protect\includegraphics[scale=0.325,valign=c]{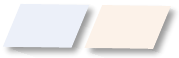} \negthinspace \negthinspace colors represent networks in initialization and inference phrases, respectively. Diagrams (d)\&(e) display comparisons of speed-precision trade-offs on CPU and Jetson AGX Xavier platforms. 
	The parameters and FLOPs are represented by the area of circles in (d) and (e), respectively.}
	\label{fig:abs}
\end{figure}

Mainstream efficient trackers can be broadly categorized into two types: siamese (two-stream) networks~\cite{lighttrack, hcat, ettrack} and one-stream networks~\cite{hit, abavitrack, mixformerv2}.
Efficient Siamese trackers build upon the success of the precision-oriented Siamese trackers~\cite{siamrpn++, transt}, where two symmetric branches with shared parameters are utilized to extract features from template and search region, respectively.
Subsequently, the designed interaction module performs feature correlation, as illustrated in Fig.~\ref{fig:abs} (a).
A typical advantage of Siamese tracking pipeline is that the detached streams allow the model to drop the template branch 
(except initialization) during inference, thereby reducing unnecessary latency.
More recently, state-of-the-art precision-oriented trackers~\cite{ostrack, mixformer, artrack,seqtrack} have evolved to universally adopt a one-stream transformer, 
\textit{beat the previous two-stream architecture}, dominating the 
tracking field.
The success of this paradigm lies in its globally receptive feature extraction and the interaction between the template and search region across the model.
Naturally, some efficient trackers follow these 
trackers and also adopt a one-stream architecture (as illustrated in Fig.~\ref{fig:abs} (b)), replacing the original computationally intensive components with lightweight backbones or modules.
For example, building on MixFormer~\cite{mixformer}, MixFormerV2~\cite{mixformerv2} proposed a one-stream fully transformer framework that utilizes distillation and transformer compression techniques, achieving real-time CPU speed.

However, the prevalent one-stream network may not be optimal for efficient visual tracking,
since 1) 
one-stream network redefines the template and search region branches as a unified structure, incorporating template computation in every video frame, which introduces significant redundancy.
2) 
the pervasive relation modeling between the template and search region overlooks the fact that this computationally intensive process imposes substantial stress on edge devices.
To address these issues, we propose a novel asymmetric Siamese tracker named \textbf{AsymTrack} for efficient tracking. 
\textit{We delve into the strengths and weaknesses of the two- and one-stream paradigms, enabling the designed AsymTrack to combine the efficiency of Siamese trackers with the precision benefits of one-stream trackers.}
As shown in Fig.~\ref{fig:abs} (c), 
AsymTrack employs two separate branches, avoiding repeated template computations during inference. The template features are transformed into modulation signals and injected into the search region branch for relation modeling.

Specifically, we design an efficient template modulation (ETM) mechanism to establish the relation modeling in two-stream architecture, which eliminates the additional interaction module as in Siamese trackers and avoids the involvement of repeated template computation as in one-stream trackers.
Furthermore, to overcome the limited representation capability in lightweight networks, we design an object perception enhancement (OPE) module.
It efficiently merges abstract semantic features and local details into the base features.
Through ingenious design, the OPE module can be flexibly re-parameterized as a single-layer convolution during inference, which improves precision while minimizing latency.
As presented in Fig.~\ref{fig:abs}, the proposed AsymTrack demonstrates excellent speed-precision trade-offs, while having fewer parameters and lower computational requirements compared to other competitors.
For instance, compared to HiT-Tiny~\cite{hit}, AsymTrack-T achieves a 6.0\% higher AUC precision on LaSOT~\cite{lasot} and exhibits faster speeds on both CPU and Jetson AGX Xavier. 
Relative to TransT~\cite{transt}, AsymTrack-B maintains a comparable AUC (64.7\% vs. 64.9\%), while operates 6.6 times faster on CPU and 4.7 times faster on AGX.
Our contributions are threefold:
\begin{itemize}
	\item
	We propose a novel asymmetric Siamese framework named AsymTrack.
	It combines the high efficiency and high precision of two- and one-stream trackers,
	surpassing the current prevailing one-stream pipeline and offering new insights for efficient tracking architecture.
	\item
	We propose an efficient template modulation (ETM) mechanism for relation modeling within our asymmetric Siamese tracking architecture, along with an object perception enhancement (OPE) module to boost the limited object representation capabilities of lightweight tracker.
	\item
	AsymTrack comprises a family of efficient trackers and
	extensive experiments demonstrate its effectiveness. Notably, the AsymTrack series, tailored for resource-constrained platforms, leads in both accuracy and speed compared to other state-of-the-art efficient trackers.
\end{itemize}

\section{Related Works}

\begin{figure*}[t]
	\centering
	\includegraphics[width=0.97\linewidth]{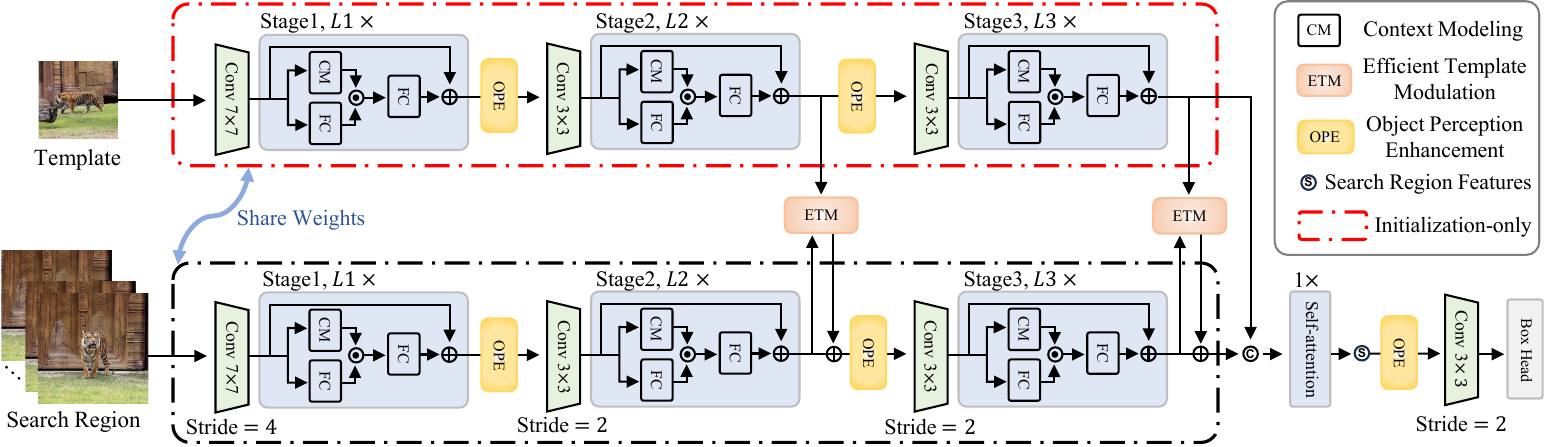}
	\caption{
		Overview of AsymTrack. 
		It employs an asymmetric Siamese pipeline, where the template branch runs once during initialization, generating features and prototype that are unidirectionally fed to the search region branch for online inference. 
	}
	\label{fig:overview}
\end{figure*}

\subsection{Precision-Oriented Tracking}
Siamese tracking pipeline~\cite{siamesefc,siamrpn++,Deeper-wider-SiamRPN,SiamBAN,transt,stark,cswintt,swintrack,donglaoshi} has long been dominant.
In these models, the template and search region are fed into two branches with shared parameters to perform feature extraction, which are then processed together through designed interaction component to achieve fusion and target object matching.
With the advent of the transformer~\cite{attention_is_all,vit}, some trackers, such as TransT~\cite{transt} and SwinTrack~\cite{swintrack}, have incorporated more powerful transformer blocks for feature fusion or representation.
Vision transformers serialize inputs into patch embeddings, allowing patches of different sizes to be concatenated along the token dimension, enabling joint modeling of the template and search region early in the model.
Consequently, a series of trackers based on a one-stream framework, such as OSTrack~\cite{ostrack}, MixFormer~\cite{mixformer}, and others~\cite{sbt, simtrack, artrack,grm}, have refreshed the state-of-the-art performance and become the mainstream paradigm for precision-oriented tracking.
Despite impressive performance, these methods are limited to desktop GPUs and often fail to meet speed requirements on resource-constrained platforms.
For example, the high-performance tracker ARTrack~\cite{artrack} runs at just 9 FPS on CPUs and 8 FPS on AGX, failing to meet basic real-time requirements.

\subsection{Efficiency-Oriented Tracking}
Efficiency-oriented tracking has recently gained attention for its potential to propel trackers toward practical applications.
The development of efficient tracking is closely tied to precision-oriented tracking. 
A common approach involves reducing the computational load of high-performance trackers by incorporating lightweight modules.
Early trackers like ECO~\cite{eco} and ATOM~\cite{atom} designed lightweight structures to reduce computational complexity in the discriminative correlation filter model~\cite{dcf}.
Following Siamese tracking pipeline, numerous Siamese efficient trackers such as LightTrack~\cite{lighttrack}, ETTrack~\cite{ettrack}, FEAR~\cite{fear}, SMAT~\cite{smat}, and LiteTrack~\cite{litetrack} emerged. 
For instance, LightTrack~\cite{lighttrack} utilized NAS~\cite{nas} to discover lightweight backbone and head networks, significantly reducing the model’s parameters and FLOPs.
To improve efficiency, 
FEAR~\cite{fear} introduced compact feature extraction and fusion blocks, 
while E.T.Track~\cite{ettrack} proposed a computationally friendly Exemplar Transformer for target localization.
As one-stream architectures like OSTrack~\cite{ostrack} and MixFormer~\cite{mixformer} have demonstrated excellent performance in precision-oriented tracking, some works~\cite{mixformerv2,hit,abavitrack} have begun exploring high-speed tracking within the one-stream framework.
Building on MixFormer~\cite{mixformer}, MixFormerV2~\cite{mixformerv2} lightens a one-stream fully transformer network through techniques such as distillation~\cite{distillation}, achieving real-time CPU speed.
HiT~\cite{hit} achieves impressive speeds on edge computing platforms by using lightweight hierarchical transformers and incorporating shallow features to compensate for information loss from large-stride downsampling.
However, despite its success in precision-oriented tracking, the one-stream architecture faces significant limitations in efficient tracking. For instance, computing the template for every video frame during inference introduces considerable redundancy. Additionally, one-stream networks rely on ViT~\cite{vit,levit,yu2024tf} backbones, where quadratic complexity relative to input size and frequent transformer attention calculations present major challenges for edge computing devices.
To this end, we delve into efficient tracker design and propose a novel asymmetric Siamese tracker that combines the strengths of both two- and one-stream trackers.

\section{Methodology}
\subsection{Preliminaries and Notation}
\noindent\textbf{{Siamese Tracking Pipeline.}}
Given the target template $\bm{Z} \in \mathbb{R}^{H_z\times W_z\times 3}$ and search region $\bm X \in \mathbb{R}^{H_x\times W_x\times 3}$, the tracker aims to estimate the object bounding box $\bm B \in \mathbb{R}^{4}$ in $\bm X$.
A typical Siamese tracker mainly consists of two symmetric feature encoder $\mathcal{F}(\cdot)$ branches with shared weights, a feature interaction module $\mathcal{I}(\cdot)$, and a box prediction head $\mathcal{\varphi}(\cdot)$.
In the initialization phase, $\bm{Z}$ is fed into $\mathcal{F}(\cdot)$ to generate the template features by $\bm H_Z=\mathcal{F}(\bm{Z})$.
In the subsequent inference phase, $\mathcal{F}(\cdot)$ extracts features from $\bm{X}$, while $\bm{H}_Z$ is directly fed into $\mathcal{I}(\cdot)$ for interaction computation and $\mathcal{\varphi}(\cdot)$ for box prediction.
The subsequent inference process can be represented as:
\begin{equation}\label{s1}
	\bm B=\mathcal{\varphi}(\mathcal{I}(\mathcal{F}(\bm X),\bm{H}_Z)).
\end{equation}

\noindent\textbf{{One-stream Tracking Pipeline.}}
Relative to Siamese trackers, one-stream trackers have a simpler architecture, typically consisting of a backbone network \(\mathcal{F}(\cdot)\) that simultaneously performs feature extraction and interaction, along with a box prediction head \(\mathcal{\varphi}(\cdot)\).
The template and search region are first converted into patches through a patch embedding layer and flattened to 1D tokens $\bm{Z}_o \in \mathbb{R}^{N_z\times D}$ and $\bm{X}_o\in \mathbb{R}^{N_x\times D}$. 
These token sequences are then concatenated along the token dimension and fed into \(\mathcal{F}(\cdot)\), which is generally a transformer network.
Leveraging the global long-range modeling of transformers, the concatenated tokens undergo extensive interaction, substantially improving the accuracy of object modeling.
The inference process can be described as:
\begin{equation}\label{s2}
	\bm B=\mathcal{\varphi}((\mathcal{F}(concat(\bm{Z}_o,\bm{X}_o)))).
\end{equation}

\subsection{Asymmetric Siamese Architecture}

\noindent\textbf{{Asymmetric Siamese Structure.}}
Siamese tracker has efficiency advantages due to the elimination of redundant template computations, but the insufficient relation modeling between the template and search region limits its performance.
One-stream tracker demonstrates superior performance, but involving the template in every frame inference and adopting dense transformer attention layer with quadratic complexity hinder its deployment on edge computing platforms.
Building on the above analysis, we propose an asymmetric Siamese tracking architecture that unites the speed of Siamese trackers with the superior performance of one-stream trackers.
To avoid redundant template computation, we first adopt a two-stream structure, using separate encoders to extract features from the template and the search region. Since the architecture is not symmetrical, we denote these encoders as \(\mathcal{F}_z\) and \(\mathcal{F}_x\), respectively. Notably, \(\mathcal{F}_z\) is a subset of the \(\mathcal{F}_x\) network, so the weights of \(\mathcal{F}_z\) are integrated into and shared with \(\mathcal{F}_x\).
To compensate for the lack of relation modeling within the two-stream structure, we propose the concept of template modulation.
During initialization, \(\mathcal{F}_z\) extracts the template features and generates template prototype by $\bm{H}_z,\bm{P}_z=\mathcal{F}_z(\bm Z)$.
$\bm{P}_z$ is then unidirectionally fed into the search region branch for modulation purposes.
This process facilitates cross-branch information communication throughout different stages of the backbone.
The inference process of the asymmetric Siamese tracking pipeline can be described as:
\begin{equation}\label{s3}
	\bm B=\mathcal{\varphi}((\mathcal{F}_x(\bm X,\bm{H}_z, \bm{P}_z))).
\end{equation}
Based on this pipeline, we present our efficient tracker AsymTrack, the overall framework is shown in Fig.~\ref{fig:overview}.

\noindent\textbf{{Lightweight Backbone.}}
We adopt a lightweight hierarchical model EfficientMod~\cite{efficientMod} as our backbone. To balance speed and accuracy, we selected the first three stages and integrated them into AsymTrack. As shown in Fig.~\ref{fig:overview}, the backbone network is duplicated into two parameter-sharing branches.
The input images first pass through a $7 \times 7$ convolutional layer for 4$\times$ downsampling, followed by a $3 \times 3$ convolutional layer after each stage for 2$\times$ downsampling.
The $i$-th stage consists of $L_i$ encoder blocks, each comprising a Context Modeling (CM) block and fully connected (FC) layers with a residual connection. The CM block, structured as FC-Conv-FC, aggregates visual contexts, while the parallel FC layer projects the input into a new space.
They are fused by element-wise multiplication, mimicking the dynamics of self-attention, followed by a linear projection after fusion. 
Following three hierarchical encoder stages, and inspired by recent advances~\cite{li2023convmlp} combining convolution and attention, we added a transformer attention layer to enhance interaction between the template and search region. 
The attention layer is introduced only after the last stage, where the feature size is relatively small.

\noindent\textbf{{Box Prediction Head.}} 
We use a simple corner head to predict the object bounding box. Following STARK~\cite{stark} but in a more streamlined way, we feed the search region features into a few stacked Conv-BN-ReLU layers to estimate the target's top-left and bottom-right coordinates. Our box head requires no extra post-processing \eg window penalty~\cite{siameserpn}.

\subsection{Efficient Template Modulation}
\begin{figure}[t]
	\centering
	\includegraphics[width=0.97\linewidth]{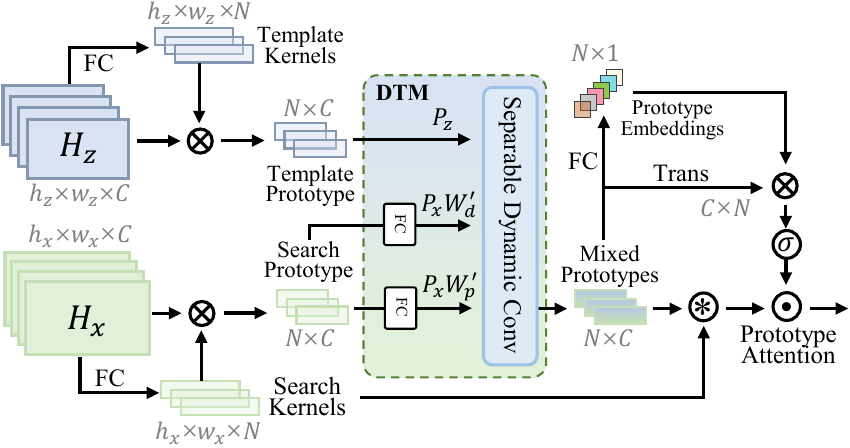}
	\caption{Efficient template modulation (ETM) mechanism. 
	}
	\label{fig:etm}
\end{figure}

Interaction between the template and search region is pivotal for accurate tracking, as demonstrated by a series of one-stream trackers~\cite{ostrack,mixformer}. However, designing an efficient interaction method in two-stream network without relying on dense and heavy transformer layers remains a challenge. To address this, we introduce an efficient template modulation mechanism, as shown in Fig.~\ref{fig:etm}.

\noindent\textbf{{Prototype Generation.}} 
We first selectively extract key information from the template and search region features $\bm{H}_z \in \mathbb{R}^{h_z\times w_z\times C}$, $\bm{H}_x \in \mathbb{R}^{h_x\times w_x\times C}$ using a linear layer $\bm{W}_k \in \mathbb{R}^{C\times N}$ and aggregate it into the corresponding kernels,
$\bm{S}_{z,x}\in \{\bm{S}_z,\bm{S}_x\}$
where $N$$<$$C$.
Next, feature contraction is performed using a dot product operation, resulting in the corresponding prototypes $\bm{P}_z, \bm{P}_x \in \mathbb{R}^{N\times C}$. This process can be expressed as:
\begin{equation}\label{s4}
	\bm{P}_{z,x}=r(\bm{H}_{z,x}*\bm{W}_k)r(\bm{H}_{z,x})^{\top},
\end{equation}
where $r(\cdot)$ denotes flattening the spatial dimension from $h\times w$ to $hw$. 
Notably, template prototype generation can be completed during initialization to save latency.

\noindent\textbf{{Dynamic Template Modulation.}}
Inspired by \cite{choromanski2020rethinking, yoso}, we utilize 1D dynamic convolution to achieve the effect of multi-head cross-attention in transformer for prototype relation modeling, making the process more lightweight.
The mixed prototypes after template modulation can be generated by:
\begin{equation}\label{s5}
	\begin{aligned}
		\bm{P}_{mix}=DyConv_{1d}(\bm{P}_z,\bm{K})&=\bm{P}_z*\bm K, \\
		\bm K&=\bm{P}_x\bm W_{c},
	\end{aligned}
\end{equation}
where $\bm W_c$ is weight to dynamically generate the kernel $\bm K$ conditioned on $\bm P_x$.
The standard convolution can be factorized into a depthwise convolution and a pointwise convolution~\cite{mobilenet}. Following this manner, we further achieve dynamic template modulation in a separable form,
\begin{equation}\label{s6}
	\begin{aligned}
		\bm P_{mix}&=SeDyConv_{1d}(\bm{P}_z, \bm K) \\
		&=(\bm{P}_z*\bm{K}\bm{W}_d)*(\bm{KW}_p) \\
		&=(\bm{P}_z*\bm{P}_x\bm{W}_{d}^{'})*(\bm{P}_x\bm{W}_{p}^{'}),
	\end{aligned}
\end{equation}
where $\bm{W}_{d}^{'}$ and $\bm{W}_{p}^{'}$ are the weights of depthwise convolution and pointwise convolution.
The modulated search cues $\bm V$ can be obtained by convolving $\bm{P}_{mix}$ with $\bm{S}_x$.

\noindent\textbf{{Prototype Attention.}} 
Finally, we designed prototype attention as a form of post attention to be applied to the modulated search features.
A linear layer is applied to compress the mixed prototype to $N\times 1$ dimension prototype embeddings $\bm{P}_{mix}^{emb}$, which represent the global information in each prototype.
We multiply $\bm{P}_{mix}^{emb}$ by the transposed prototype $\bm{P}_{mix}^{'}$ to generate an attention vector in $C\times 1$ dimension, which where be weighted to output modulated cues in channel dimension to emphasize important features and enhance representation.
The calculation of the whole process is:
\begin{equation}\label{s7}
	\bm{V}^{'}=\sigma(\bm{P}_{mix}^{'}\bm{P}_{mix}^{emb})\odot \bm V,
\end{equation}
where $\odot$ represents the broadcast Hadamard product and $\sigma(\cdot)$ indicates the Sigmoid function.

\begin{figure}[t]
	\centering
	\includegraphics[width=0.925\linewidth]{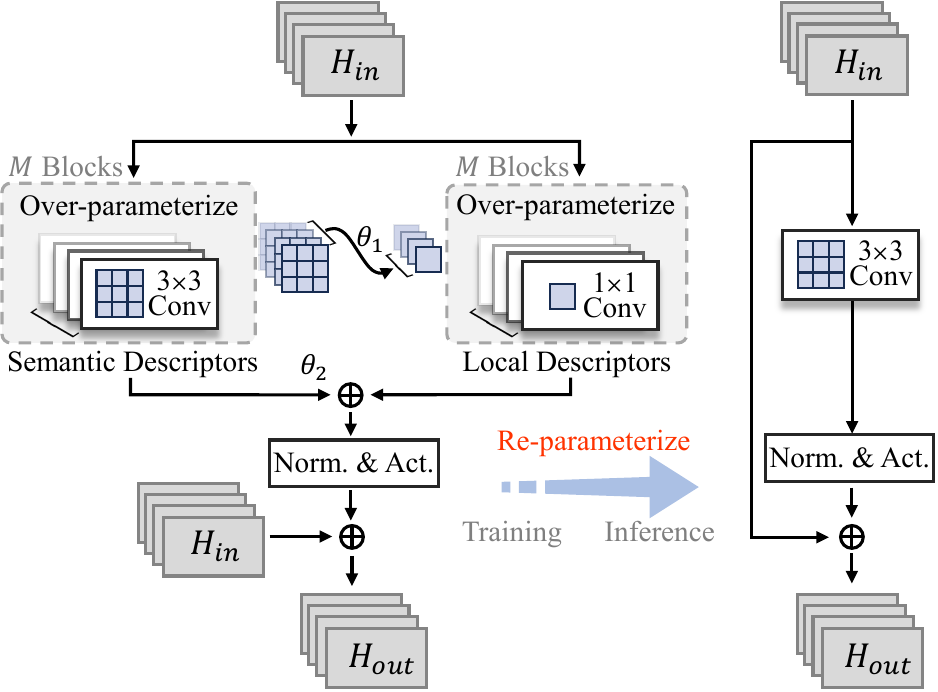}
	\caption{
		Object perception enhancement (OPE) module.
	}
	\label{fig:ope}
\end{figure}

\subsection{Object Perception Enhancement}
\noindent\textbf{{Joint Semantic and Local Representation.}} 
To tackle the bottleneck of light-weight tracker's representation optimization, we propose an object perception enhancement (OPE) module, detailed in Fig.~\ref{fig:ope}.
High-level task encourages standard convolution to consolidate the abstract semantics~\cite{li2010object} and local cues can be extracted from gradient information~\cite{su2021pixel},
OPE enhances backbone features by combining the representations of semantic and local descriptors, whch are achieved by $3\times 3$ and $1\times 1$ convolutional kernels, respectively.
Specifically, the local descriptor is a parameter reusing mechanism generated by the linear transformation of the integrating gradient cues around shared semantic descriptors.
The detail cues \eg object boundary and texture cues can be captured by the local descriptors.
Denote a single semantic descriptor as $\bm W_{sem}\in\mathbb{R}^{C_{out}\times C_{in}\times 3\times 3}$, where $C_{out}$/$C_{in}$ is the output/input channel dimension, a local descriptor $\bm{W}_{loc}$ can be generated by,
\begin{equation}\label{s8}
	\bm{W}_{loc}=-\theta_{1}\sum_{p_n\in 
		\mathbb{S}}\bm{W}_{sem}(p_n),
\end{equation} 
where $p_n$ is the $n$-th weight value in $3\times 3$ convolution $\mathbb{S}$. 
$\theta_{1}$ is a learnable parameter projection factor.
The enhanced representation is obtained by linearly weighting the extracted local features and semantic features with a weight \(\theta_2\).
Furthermore, we utilize the over-parameterization technique~\cite{guo2020expandnets} to further enhance perceptual performance. The semantic descriptor branch and the local descriptor branch are implemented as $M$ parallel branches.

\noindent\textbf{{Re-parameterization Inference.}} 
For inference, we leverage the homogeneity and additivity of the convolutions to fold semantic and local descriptors into a single $3\times 3$ convolution, and convert over-parameterization parallel branches into an equivalent single branch through a linear transformation $\bm W = \sum_{i=1}^{M}\bm{W}_i$, without performance degradation.

\subsection{Optimization and Tracker Inference}
\noindent\textbf{{Optimization.}} 
The training objective consists of a $\mathcal{L}_1$ loss and GIoU loss~\cite{giou} $\mathcal{L}_G$, 
\begin{equation}\label{eq:loss}
	\mathcal{L}=\lambda_{1}\mathcal{L}_1(\bm B,\bm{B}_{gt})+\lambda_G\mathcal{L}_G(\bm B,\bm{B}_{gt}),
\end{equation} 
where $\bm{B}_{gt}$ is the ground truth, 
$\lambda_1=5$ and $\lambda_G=2$.

\noindent\textbf{{Tracker Inference.}} 
AsymTrack features an asymmetric Siamese structure. The template branch runs only once during initialization, after which only the search region branch is needed for inference, with the template's extracted features and modulation cues injected.

\begin{table}[!h]
	\centering
	\resizebox{0.475\textwidth}{!}{
		\begin{tabular}{c|c|c|c|c}
			\Xhline{1.2pt}
			\multicolumn{2}{c|}{Model} & {AsymTrack-T} & {AsymTrack-S} & {AsymTrack-B} 
			\\
			\hline
			\multicolumn{2}{c|}{Encoder Blocks} & $\left[2,2,1\right]$ &$\left[2,2,3\right]$ & $\left[2,2,3\right]$ 
			\\
			\hline
			\multicolumn{2}{c|}{Input Sizes} & $\left[128,256\right]$ & $\left[128,256\right]$ & $\left[192,384\right]$ \\
			\hline
			\multirow{3}{*}{\makecell{Inference Speed\\(FPS)}} &{GPU} & 224 & 200 & 197 \\
			&{CPU} & 81 & 75 & 38 \\
			&{AGX} & 84 & 78 & 64 \\
			\hline
			\multicolumn{2}{c|}{Params (M)} &3.05 &3.36 &3.36 \\
			\hline
			\multicolumn{2}{c|}{FLOPs (G)} &0.7 &0.8 &1.8 
			\\
			\Xhline{1.2pt}
		\end{tabular}}
	\caption{Detailed configurations of our AsymTrack variants.}
	\label{tab:config}
\end{table}

\section{Experiments}
\label{sec:Experiments}
\subsection{Implementation Details}

\textbf{{AsymTrack Model Family.}} 
We present three variants of the AsymTrack model: AsymTrack-T, AsymTrack-S, and AsymTrack-B. Tab.~\ref{tab:config} details their configurations, also including parameters, FLOPs, and inference speeds across different platforms: GPU (Nvidia 2080ti), CPU (Intel i7-9700KF@3.6G Hz), and edge device (Jetson AGX Xavier).

\noindent\textbf{{Training Details.}} We used training splits of four datasets for training, including LaSOT~\cite{lasot}, TrackingNet~\cite{trackingnet}, COCO2017~\cite{coco}, and GOT10K~\cite{got10k}. 
Common augmentation such as flipping and jittering are applied.
The template and search region images are resized to 128 $\times$ 128 and 256 $\times$ 256 for AsymTrack-T and AsymTrack-S, and to 192 $\times$ 192 and 384 $\times$ 384 for AsymTrack-B. We trained the model for 500 epochs using the AdamW~\cite{adamw} optimizer with an initial learning rate of 4e-4 on 2 NVIDIA A800 GPUs, with each epoch consisting of 60,000 randomly sampled image pairs.

\begin{table*}[t]
	\footnotesize
	\centering
	\renewcommand\arraystretch{1.}
		\resizebox{\textwidth}{!}{
			\begin{tabular}{c|l|ccc c ccc c ccc c ccc}
				\Xhline{1.2pt}
				& \multirow{2}*{Method} &  \multicolumn{3}{c}{GOT-10k}&& \multicolumn{3}{c}{LaSOT} &&  \multicolumn{3}{c}{TrackingNet} && \multicolumn{3}{c}{PyTorch Speed (\emph{fps})}\\
				\cline{3-5}
				\cline{7-9}
				\cline{11-13}
				\cline{15-17}
				&& AO&SR$_{0.5}$&SR$_{0.75}$ && AUC&P$_{Norm}$&P && AUC&P$_{Norm}$&P && GPU&CPU&AGX\\
				\hline
				\multirow{14}*{\rotatebox{90}{Real-time}}
				&AsymTrack-B (ours) 
				
				&\textbf{{67.7}}&\textbf{{76.6}}&\textbf{{61.4}} 
				
				&&\textbf{{64.7}}&{\underline{73.0}}&{\underline{67.8}}
				
				&&\textbf{{80.0}}&\textbf{{84.5}}&\textbf{{77.4}}
				
				&&197&38&{64}\\
				
				&AsymTrack-S (ours) 
				
				&65.5 &74.8 &58.9
				&&62.8 &71.2 &64.8
				&&77.9&{82.2}&74.0  &&200&\textbf{75}&\textbf{78}\\
				
				&AsymTrack-T (ours)
				&62.3 &71.3 &54.7
				&&60.8 &68.7 &61.2
				&&76.2&80.9&71.6 &&\textbf{224}&\textbf{81}&\textbf{84}\\
				
				&HiT-Base~\cite{hit}& 64.0 & 72.1 &{58.1}&& {\underline{64.6}} & \textbf{{73.3}} & \textbf{{68.1}}&& \textbf{{80.0}}&{\underline{84.4}}&{\underline{77.3}} && 175&33&61\\
				&TCTrack~\cite{cao2022tctrack}& {\underline{66.2}}&75.6&{\underline{61.0}} && 60.5&69.3&62.4 && 74.8&79.6&{73.3} && 140&45&41\\
				&MixFormerV2~\cite{mixformerv2}& 61.9&71.7&51.3 && 60.6&69.9&60.4 && 75.8&81.1&70.4 && 167&45&49\\
				&HCAT~\cite{hcat}& {65.1}&{\underline{76.5}}&{56.7} && {59.3}&{68.7}&{61.0} &&  {76.6}&82.6&{72.9} &&195&45&55 \\
				&HiT-Small~\cite{hit}& 62.6 & 71.2 & 54.4 && 60.5 & 68.3 & 61.5&& {77.7}&{81.9}&73.1 && 192&72&68\\
				&E.T.Track~\cite{ettrack}& -&-&- && 59.1&-&- &&  75.0&80.3&70.6 &&40&47&20\\
				&FEAR~\cite{fear}& 61.9&{72.2}&- && 53.5&-&54.5 &&  -&-&- && 105&60&38\\
				&LightTrack~\cite{lighttrack}& 61.1&71.0&- && 53.8&-&53.7 &&  72.5&77.8&69.5 && 128&41&36\\
				&ATOM~\cite{atom}& 55.6&63.4&40.2 && 51.5&57.6&50.5 &&  70.3&77.1&64.8 &&83&18&22 \\
				&HiT-Tiny~\cite{hit}& 52.6 & 59.3 & 42.7 && 54.8 & 60.5 & 52.9&& 74.6&78.1&68.8 && \textbf{204}&\textbf{76}&\textbf{77}\\
				&ECO~\cite{eco}& 31.6&30.9&11.1 && 32.4&33.8&30.1 &&  55.4&61.8&49.2 && \textbf{240}&15&39 \\
				\hline
				\multirow{11}*{\rotatebox{90}{Non-real-time}}
				&ARTrack~\cite{artrack}& 73.5&82.2&70.9 && 70.4&79.5&76.6 && 84.2&88.7&83.5 && 26&9&8\\
				&MixFormer-L~\cite{mixformer}& {75.6}&{85.7}&{72.8} && {70.1}&{79.9}&{76.3} &&  {83.9}&{88.9}&83.1 &&18&-&-\\
				&TransT~\cite{transt}& 72.3&82.4&68.2 && 64.9&73.8&69.0 && 81.4&86.7&80.3 && 63&5&13  \\
				&OSTrack-256~\cite{ostrack}& 71.0&80.4&68.2 && 69.1&78.7&75.2 &&  83.1&87.8&82.0 && 105&11&19 \\
				&Sim-B/16~\cite{simtrack}& 68.6&78.9&62.4 && 69.3&78.5&- && 82.3&-&{86.5} && 87&10&16 \\
				&STARK-ST50~\cite{stark}& 68.0&77.7&62.3  && 66.6&-&- && 81.3&86.1&- && 50&7&13 \\
				&TrSiam~\cite{trdimp}& 67.3&78.7&58.6 && 62.4&-&60.6 &&  78.1&82.9&72.7 && 40&5&10 \\
				&DiMP~\cite{DiMP}& 61.1&71.7&49.2 && 56.9&65.0&56.7 && 74.0&80.1&68.7 && 77 &10&17  \\
				&SiamFC++~\cite{xu2020siamfc++}& 59.5&69.5&47.9 && 54.5&-&54.7 && 75.4&80.0&70.5 && -&12&-\\
				&SiamRPN++~\cite{siamrpn++}& 51.7&61.6&32.5 && 49.6&56.9&49.1 && 73.3&80.0&69.4 && 56 &4&11 \\
				\Xhline{1.2pt}
			\end{tabular}
	}
	\caption{
	State-of-the-art comparison on the TrackingNet, LaSOT, and GOT-10k benchmarks. The top two real-time results are highlighted in \textbf{bold} and \underline{underlined}, respectively. The top three speed across different platforms are highlighted in \textbf{bold}. }
	\label{tab-sota}
\end{table*}

\subsection{Comparison with State-of-the-arts} 
We conduct a comprehensive comparison across seven widely used benchmarks. Trackers are categorized as either real-time or non-real-time based on their speed (20 FPS) on the Jetson AGX Xavier, in accordance with the VOT~\cite{vot21} real-time criteria.

\noindent\textbf{{GOT-10k.}} 
GOT-10k~\cite{got10k} is a large-scale tracking dataset with over 10,000 video sequences featuring diverse objects and scenes. As shown in Tab.~\ref{tab-sota}, AsymTrack-B achieves the highest real-time AO score of 67.7\%.
It surpasses the recent efficient tracker HiT-Base~\cite{hit} by a large margin of 3.7\% while achieving faster speeds across all test platforms. Besides, our fastest variant, AsymTrack-T, stands out as the fastest among all real-time trackers, with comparable precision.

\noindent\textbf{{LaSOT.}} 
LaSOT~\cite{lasot} is a large-scale long-term benchmark comprising 1,400 video sequences, each averaging over 2,500 frames, with 280 sequences reserved for testing.
As shown in Tab.~\ref{tab-sota}, AsymTrack-B achieved the best AUC score of 64.7\% and outperformed the previous state-of-the-art HiT-Base~\cite{hit} in speed across all platforms. While AsymTrack-T leads in speed on the AGX (84 FPS), it ranks fourth among real-time trackers. Notably, it outperformed HiT-Tiny by 6.0\% in AUC, 8.2\% in normalized precision, and 6.7\% in precision.

\noindent\textbf{{TrackingNet.}} 
TrackingNet~\cite{trackingnet} is a large-scale short-term benchmark with 511 test video sequences, covering a wide range of object categories and scenes.
As shown in Tab.~\ref{tab-sota}, AsymTrack-B delivered top-tier performance with an AUC of 80.0\%, normalized Precision of 84.5\%, and precision of 77.4\%. The smaller variant, AsymTrack-S, is also highly competitive, surpassing the latest one-stream tracker MixFormerV2~\cite{mixformerv2} by 2.1\% in AUC and 3.6\% in precision.

\noindent\textbf{{Speed Comparison.}} 
We conducted speed tests across three different platforms. As we can see, ECO~\cite{eco} and our AsymTrack-T are the two fastest trackers, with AsymTrack-T achieving 224 FPS on GPU, 81 FPS on CPU, and 84 FPS on AGX. On resource-constrained platforms like the CPU and AGX, AsymTrack-T outperforms ECO in speed while far surpassing it in precision. 
The smaller variant, AsymTrack-S, is 1.2$\times$ faster on GPU, 1.7$\times$ faster on CPU, and 1.6$\times$ faster on AGX compared to the recent one-stream tracker MixFormerV2-S~\cite{mixformerv2}. 
The base variant, AsymTrack-B, is 22 FPS faster on GPU, 5 FPS faster on CPU, and 3 FPS faster on AGX compared to its counterpart, HiT-Base~\cite{hit}. 
In summary, AsymTrack delivers impressive speed across multiple platforms, making it highly suitable for application scenarios \eg UAVs and embodied robots.

\noindent\textbf{{NFS.}} 
NFS~\cite{NFS} is a high-frame-rate dataset focused on fast-motion object scenarios. As shown in Tab.~\ref{tab-sota-small},  on 30 FPS version of NFS, our tracker excels in these challenging conditions, achieving the top two AUC scores.

\noindent\textbf{{UAV123.}} 
UAV123~\cite{uav} aims to focus on the challenges unique to UAV-based tracking.
AsymTrack-B achieves the highest AUC score of 66.5\% As shown in Tab.~\ref{tab-sota-small}, outperforming 
MixFormerV2~\cite{hit} by 0.7\% and HiT-Base~\cite{hit} by 0.9\%.

\noindent\textbf{{LaSOT$_{ext}$.}} 
LaSOT$_{ext}$~\cite{lasotext}, an extension of LaSOT for more challenging tracking evaluations, further demonstrates the strength of our approach.
AsymTrack-B ranks first with an AUC of 44.6\%, and AsymTrack-T outperforms HiT-Small by 2.1\% with a speed advantage.

\begin{table}[h]
	\small
	\centering
	\renewcommand\arraystretch{1.}
		\resizebox{\linewidth}{!}{
			\begin{tabular}{c|l|ccc}
				\Xhline{1.2pt}
				&Method&NFS&UAV123&LaSOT$_{ext}$\\
				\hline
				\multirow{13}*{\rotatebox{90}{Real-time}}
				&AsymTrack-B (ours) & {\underline{64.4}} & \textbf{{66.5}} & \textbf{{44.6}}\\
				&AsymTrack-S (ours)& \textbf{{64.9}} & 65.6 & {43.3}\\
				&AsymTrack-T (ours)& 63.3 & 64.6 & 42.5\\
				&MixFormerV2~\cite{mixformerv2} & - & {\underline{65.8}} & 43.6\\
				&HiT-Base~\cite{hit} &63.6 &65.6 &{\underline{44.1}}\\
				&HCAT~\cite{hcat}&{63.5}&62.7&-\\
				&HiT-Small~\cite{hit} & {61.8} & {63.3} & {40.4}\\
				&E.T.Track~\cite{ettrack}&59.0&62.3&-\\
				&FEAR~\cite{fear}&61.4&-&-\\
				&ATOM~\cite{atom}&58.4&{64.2}&{37.6}\\ 
				&LightTrack~\cite{lighttrack}&55.3&62.5&-\\
				&HiT-Tiny~\cite{hit} &53.2 &58.7 &35.8\\
				&ECO~\cite{eco}&46.6&53.2&22.0\\
				\hline
				\multirow{8}*{\rotatebox{90}{Non-real-time}}
				&GRM~\cite{sun2023grm}&65.6&70.2&-\\
				&ARTrack~\cite{artrack}&64.3&67.7&46.4\\
				&OSTrack-256~\cite{ostrack}&64.7&68.3&47.4\\
				&TrSiam~\cite{trdimp}&65.8&67.4&-\\
				&TransT~\cite{transt}&65.7&69.1&-\\
				&PrDiMP~(Danelljan et al. 2020) &63.5&68.0&-\\ 
				&DiMP~\cite{DiMP} &62.0&65.3&39.2\\
				&SiamRPN++~\cite{siamrpn++}&50.2&61.6&34.0\\
				\Xhline{1.2pt}
			\end{tabular}
	}
	\caption{State-of-the-art comparison on more benchmarks.}
	\label{tab-sota-small}
\end{table}

\begin{figure}[t]
	\centering
	\includegraphics[width=0.995\linewidth]{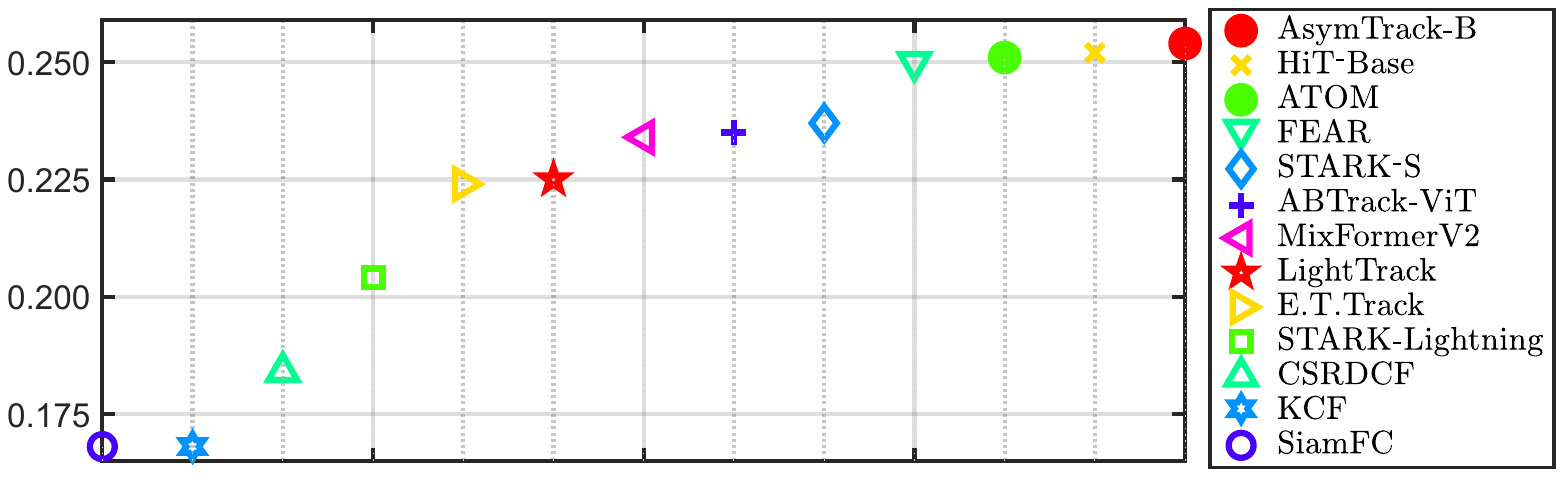}
	\caption{VOT real-time testing on Jetson AGX Xavier.}
	\label{fig:vot21}
\end{figure}

\noindent\textbf{{VOT2021.}} We also conducted real-time experiments on VOT2021~\cite{vot21} challenge benchmark running on the edge device of Jetson AGX Xavier. 
As shown in Fig.\ref{fig:vot21}, AsymTrack-B achieves 25.4\% in terms of EAO score, surpassing all other real-time efficient trackers.

\subsection{Exploration Studies}

We further conduct experiments to explore the characteristics of our AsymTrack.
LaSOT~\cite{lasot} and GOT-10k~\cite{got10k} are employed as the evaluation datasets.
AsymTrack-S is used as the baseline model by default, and unless otherwise stated, the experimental settings are kept the same as the baseline.

\begin{figure}[t]
	\centering
	\includegraphics[width=0.95\linewidth]{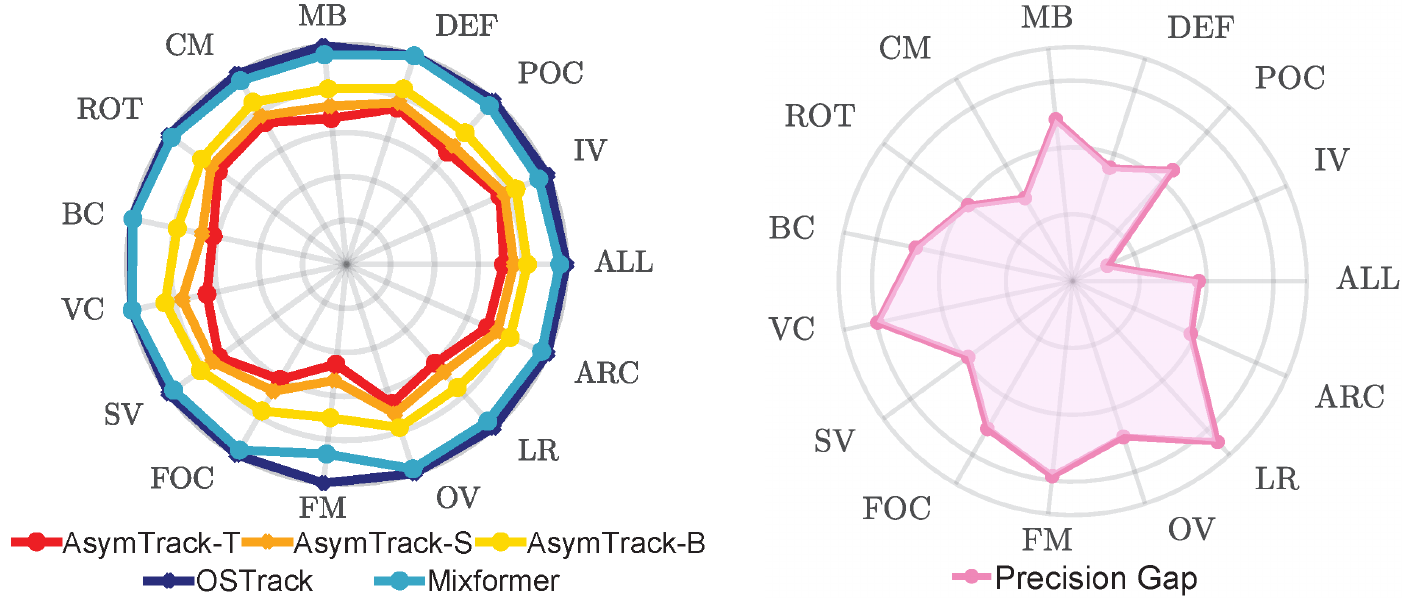}
	\caption{Gap analysis between AsymTrack and precision-oriented trackers across different attributes on LaSOT.}
	\label{fig:gap}
\end{figure}

\begin{figure}[t]
	\centering
	\includegraphics[width=0.95\linewidth]{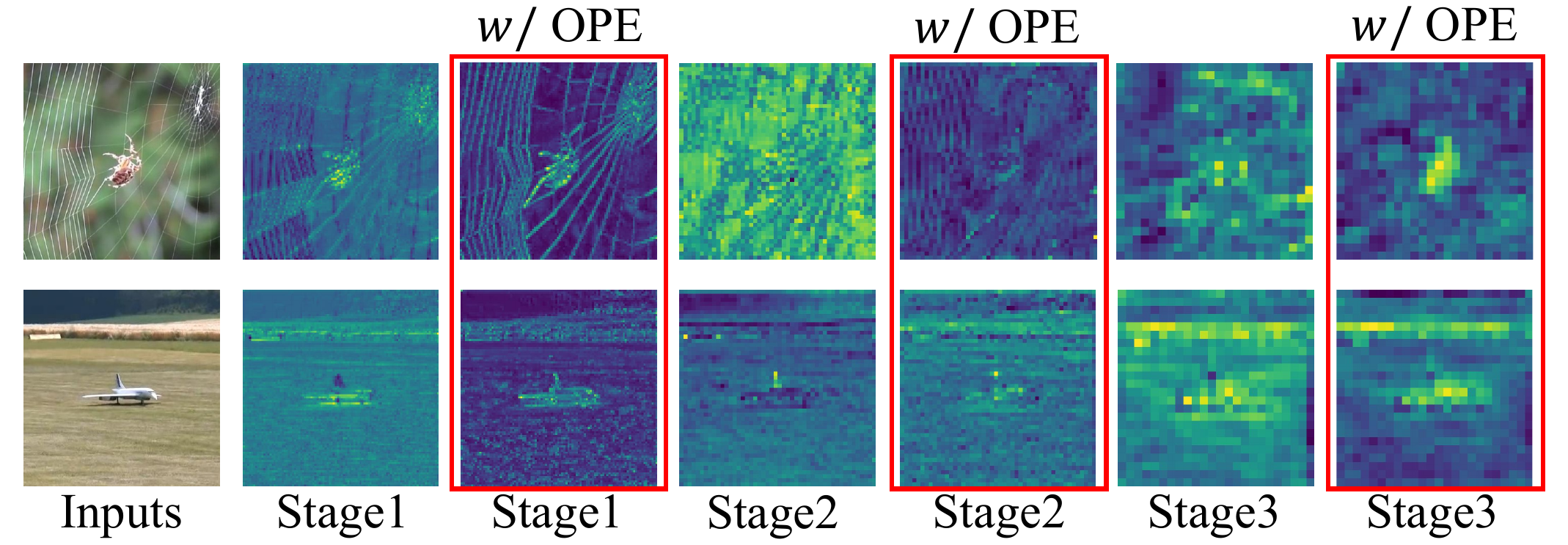}
	\caption{
		Visualization maps between $w/o$ and $w/$ OPE.
	}
	\label{fig:ope_vis}
\end{figure}

\noindent\textbf{{Gap Analysis with Precision-Oriented Trackers.}} 
We compare AsymTrack with precision-oriented trackers and, as shown in Fig.~\ref{fig:gap}, there is still a significant performance gap between AsymTrack and alternatives like OSTrack~\cite{ostrack}.
The right graph shows the average AUC gap between AsymTrack variants and precision-oriented trackers across 14 attributes. The largest gaps appear in low resolution, viewpoint change, and fast motion, where the model's representation capability is more challenged. We hope future designs for efficient tracking will further narrow the gap.

\begin{table}[t]
	\setlength{\tabcolsep}{7pt}
	\begin{center}
		\resizebox{0.45\textwidth}{!}{
			\begin{tabular}{cccccccc}
				\Xhline{1.2pt}
				\multirow{2}{*}{Model} & \multicolumn{2}{c}{Components} & \multicolumn{3}{c}{Model Speed (FPS)}& \multirow{2}{*}{GOT-10k}& \multirow{2}{*}{LaSOT} \\
				& ETM & OPE  &GPU &CPU &AGX & &  \\ \hline
				1                       & \ding{56}   & \ding{56}   &260&95&100&60.4&58.8    \\
				2&\ding{52}& \ding{56}&220&80&82&62.9&61.2\\
				3&\ding{56}&\ding{52}&240&86&92&62.1&60.7\\
				4&\ding{52}&\ding{52}&200&75&78&64.4&62.5\\ 
				\Xhline{1.2pt}
			\end{tabular}
		}
	\end{center}
	\caption{Component-wise study on performance and speed.}
	\label{table:abla}
\end{table}

\noindent\textbf{{Necessity of Template Modulation.}}
To validate the effectiveness of template modulation, we compare AsymTrack with models without this feature and evaluate their performance and speed. In Tab.~\ref{table:abla}, model\#2 and model\#4 outperform model\#1 with AUC gains of 2.5\%/4.0\% on GOT-10k and 2.4\%/3.7\% on LaSOT, confirming the benefits of template modulation. Moreover, the speed remains high across all platforms, reflecting a good speed-precision trade-off.

\noindent\textbf{{Effectiveness of Perception Enhancement.}}
To further validate the effectiveness of OPE, we conducted another comparative analysis between AsymTrack and models lacking this enhancement.
The results in Tab.~\ref{table:abla} (model\#1 and model\#3) show that incorporating OPE improves the AO on GOT-10k by 1.7\% and the AUC on LaSOT by 1.9\%. More importantly, thanks to the introduction of re-parameterization inference, the OPE module adds minimal latency across different platforms. 
Fig.~\ref{fig:ope_vis} also demonstrates that with our OPE, more discriminative features, particularly crucial detail cues, are obtained across different stages.

\noindent{\textbf{Ablation on ETM Designs.}}
As shown in Tab.~\ref{tab:study} (a), we explore different ETM designs to validate its effectiveness. Replacing our DTM with vanilla multi-head cross attention (MHCA) yields baseline-level performance but with lower efficiency. Prototype attention in ETM also proves effective. Adding ETM in early stage1 brings little improvement, suggesting early interaction has limited benefit.

\noindent{\textbf{Location Analysis of OPE.}}
We further investigate the effect of placing OPE at different stages of the model. As shown in Tab.~\ref{tab:study} (b), applying perception enhancement in the early stages is more effective than doing so at later stages, and applying it at every stage yields the best results.

\begin{table}[h]
	\centering
	\setlength{\tabcolsep}{5pt}
	\begin{subtable}[t]{0.2325\textwidth}
		\footnotesize
		\centering
		\resizebox{1\linewidth}{!}{
			\begin{tabular}{c|cc}
				\Xhline{1.2pt}
				Method & GOT-10k & LaSOT \\
				\hline
				\textit{Baseline} & 64.4 & 62.5  \\
				\textit{DTM}$\rightarrow$\textit{MHCA}& 64.1 & 62.6 \\
				$-$ \textit{Prototype Att} & 63.9 & 62.3 \\
				$+$ \textit{Stage1 ETM} & 64.6 & 62.4 \\
				\Xhline{1.2pt}
		     \end{tabular}}
	     \caption{Ablation on ETM designs}
	\end{subtable}
	\hfill
	\begin{subtable}[t]{0.2325\textwidth}
		\footnotesize
		\centering
		\resizebox{1\linewidth}{!}{
			\begin{tabular}{c|cc}
				\Xhline{1.2pt}
				Method & GOT-10k & LaSOT \\
				\hline
				\ \ \ \ \ \ \textit{Baseline} \ \ \ \ \ \ & 64.4 & 62.5  \\
				\textit{\{s1,s2\}} &64.2	&62.2 \\
				\textit{\{s1\}} &63.9	&61.9 \\
				\textit{\{s3\}} &63.3	&61.4 \\
				\Xhline{1.2pt}
		\end{tabular}}
	\caption{Different locations for OPE}
	\end{subtable}
	\caption{Ablation Study of ETM designs and OPE location.}
	\label{tab:study}
\end{table}

\section{Conclusion}
In this work, we present AsymTrack, a new family of efficient tracking models. Departing from the prevalent one-stream architectures, AsymTrack utilizes a novel asymmetric Siamese framework that integrates the efficiency of two-stream trackers with the performance benefits of one-stream designs. 
AsymTrack broadens the possibilities for real-time visual tracking on resource-constrained platforms, offering a viable alternative to one-stream architectures. We envision the AsymTrack family becoming a dependable visual tracking solution for real-world deployment, bridging the gap between academic research and industrial applications.

\section{Acknowledgments}
The paper is supported by the National Natural Science Foundation of China under grant No. 62293540, 62293542, 62106149, Liao Ning Province Science and Technology Plan No.2023JH26/10200016 and Dalian City Science and Technology Innovation Fund No. 2023JJ11CG001.

\bibliography{vot}


\end{document}